%% file: main.tex
\RequirePackage{iftex}
\newcommand{\SIAMXeTeXCompatibility}{%
  \AddToHook{package/breakurl/before}{%
    \let\SIAMSavedIfPDF\ifpdf
    \let\ifpdf\iftrue}%
  \AddToHook{package/breakurl/after}{%
    \let\ifpdf\SIAMSavedIfPDF}%
}
\ifXeTeX\SIAMXeTeXCompatibility\fi

\documentclass[hidelinks,nohypdvips]{siamart251216}

\usepackage{amsfonts,amssymb}
\usepackage[nopatch=eqnum]{microtype}
\usepackage[american]{circuitikz}
\usepackage{booktabs}
\usepackage{tabularx}

\newcommand{\rev}[1]{\textcolor{blue}{#1}}

\crefformat{equation}{\rev{Equation}~\textup{#2(#1)#3}}
\Crefformat{equation}{\rev{Equation}~\textup{#2(#1)#3}}
\crefrangeformat{equation}{\rev{Equations}~\textup{#3(#1)#4--#5(#2)#6}}
\Crefrangeformat{equation}{\rev{Equations}~\textup{#3(#1)#4--#5(#2)#6}}
\crefmultiformat{equation}{\rev{Equations}~\textup{#2(#1)#3}}%
  { and~\textup{#2(#1)#3}}{, \textup{#2(#1)#3}}{, and~\textup{#2(#1)#3}}
\Crefmultiformat{equation}{\rev{Equations}~\textup{#2(#1)#3}}%
  { and~\textup{#2(#1)#3}}{, \textup{#2(#1)#3}}{, and~\textup{#2(#1)#3}}

\definecolor{inputfill}{HTML}{D6E4F5}\definecolor{inputdraw}{HTML}{4A78B0}
\definecolor{mediumfill}{HTML}{DCEFD9}\definecolor{mediumdraw}{HTML}{4C9A5B}
\definecolor{centerfill}{HTML}{FFF2C7}\definecolor{centerdraw}{HTML}{B8932E}
\definecolor{readfill}{HTML}{FCE1CC}\definecolor{readdraw}{HTML}{CC7A3A}
\definecolor{outfill}{HTML}{E7DEF3}\definecolor{outdraw}{HTML}{7E62A8}

\title{Designing Nature Processing Units: Physical Computing Devices}
\author{Eldad Haber, Eshed Gal, Moshe Eliasof, Lior Horesh}
\headers{Designing Nature Processing Units}{E. Haber}
\hypersetup{
  pdftitle={Designing Nature Processing Units: Physical Computing Devices},
  pdfauthor={Eldad Haber, Eshed Gal, Moshe Eliasof, Lior Horesh},
  pdfsubject={Introduction draft on the design of physical computing devices}
}


\begin{document}
\maketitle

\begin{abstract}
We develop a framework for designing passive physical systems that perform
learned computations through their natural dynamics. Electrical circuits
provide concrete realizations: nonlinear resistive networks compute through
equilibrium, resistor-capacitor networks through diffusion, and inductive
networks through wave propagation and physical memory. Their component values, input couplings, and readouts are trained on chosen
connectivities subject to passivity and realizability constraints. We illustrate the approach with three tasks: static
classification using a bipartite resistive network, image denoising using
parallel nonlinear diffusion channels, and character-level language modeling
using a two-sided three-layer inductive medium trained by distillation from a
transformer. The experiments illustrate task-motivated physical architectures in which
useful computation arises from dynamics already native to the device. 
These results
motivate hardware-algorithm co-design in which the governing equations form
part of the learned architecture.
\end{abstract}

\begin{keywords}
physical computing devices, partial differential equations, passive systems,
inverse design, neural networks, energy efficiency
\end{keywords}

\begin{MSCcodes}
68T07, 65K10
\end{MSCcodes}

\section{Introduction}
\label{sec:introduction}

Neural networks have become the dominant computational engines for tasks such as
image and language processing. Their success, however, comes with substantial
energy demands that limit their scalability and
deployment~\cite{sze2017efficient}. Our goal in this study is to put forward a framework for the  design of physical
computing devices that perform such tasks at a substantially lower energy cost.

The energy cost depends on both the computation and its implementation.
Conventional neural networks evaluate learned transformations through sequences
of matrix operations and nonlinear activations. In addition to arithmetic,
these computations require memory access, movement of parameters and
intermediate states, and communication between computational units. Reducing
the energy of individual multiply-accumulate (MAC) operations therefore
addresses only part of the cost~\cite{horowitz2014energy,sze2017efficient}.
This motivates us to consider devices in which a physical medium performs a
learned transformation through its response to an input.

Our starting point capitalizes on the connection between neural networks and differential
equations. In 2017, we and others interpreted deep residual networks as
discretizations of ordinary differential equations, allowing network
architecture, stability, and learning to be studied using tools from dynamical
systems~\cite{haber2018stable,e2017proposal}. Subsequent work on neural ordinary
differential equations developed continuous representations of network depth
and methods for training them~\cite{chen2018neuralode}.

Additionally, residual networks can be related to spatial differential operators and to
parabolic and hyperbolic equations~\cite{ruthotto2020pde}. Similarly, graph neural networks
admit similar interpretations through diffusion, transport, and reaction on
discrete domains~\cite{eliasof2021pdegcn,eliasof2024graph,eliasof2024transport}.
These interpretations associate network operations with familiar physical
processes: diffusion mixes information, transport moves it, and reactions
transform it locally. Wave propagation provides a mechanism for transmitting
information and retaining a temporal state.

In this work, we reverse this line of reasoning and understanding of neural networks. That is, rather than interpreting neural networks as physical processes, we use differential equations
to \emph{design} physical devices that carry out a desired computation. We refer to
these devices as \emph{nature processing units} (NPUs). The central idea is to replace
sequences of MAC operations with the response of a physical system. The input
excites the device, interactions between its components determine an internal
state, and observations of that state provide the output. The physical process
itself performs the computation, whether through relaxation to equilibrium,
wave propagation, or diffusion and reaction within a material. A response that
requires many arithmetic operations to simulate digitally can occur directly
in a device. The main challenge is to find a realizable design whose response
performs the desired task at a low energy cost.

\textbf{Related works.}
The design of a media to match physical response (input-output) is the cornerstone on imaging \cite{dcresistivity, eit, seismicimaging}. Using techniques such as the adjoint to compute derivatives exist since the 70's \cite{lions1971optimal} and are now fairly well understood \cite{haber2000optimization, plessix2006adjoint}.

The Use of these techniques to the design of physical systems is, in its infancy.
Hughes et al.~\cite{hughes2019wave} relate wave dynamics to recurrent neural
networks and optimize an inhomogeneous medium by differentiating through the
wave equation, equivalently using its adjoint. Wright et
al.~\cite{wright2022physical} train sequences of physical transformations
using physical forward passes and model-based backward passes. These works
provide close precedents for learning with native physical responses and
for using evolving physical states in computation.

Hermans et al.~\cite{hermans2015trainable} used reciprocity to
backpropagate errors through physical dynamics, including an acoustic
experiment with trained input and output encodings around a fixed medium
with nonlinear feedback. Furuhata et al.~\cite{furuhata2021physical}
formulated learning in continuous-time dynamical systems as an
adjoint-based optimal-control problem and demonstrated it numerically
with an optoelectronic delay-system model.

Physical reservoir computing also exploits native dynamics for
temporal processing, typically keeping the reservoir fixed while training
its readout~\cite{tanaka2019reservoir}. Our design variables include the
physically admissible component laws that govern the medium's response.

Our recent study of static passive nonlinear media
uses nonlinear equilibria to define learned transformations. Kendall et al.~\cite{kendall2020training} formulated analog
neural computation and training through the electrical equilibria of nonlinear
resistive networks, while Scellier et al.~\cite{scellier2022agnostic} developed
agnostic equilibrium propagation for energy-minimizing physical systems
with controllable parameters, without requiring explicit derivatives or
detailed knowledge of the system.

Stern et al.~\cite{stern2021supervised} developed coupled learning
rules for flow and elastic networks by comparing free and clamped
responses. Dillavou et al.~\cite{dillavou2022demonstration} demonstrated
local learning with this approach in twin networks of variable resistors.

In our setting, training is performed in simulation using the adjoint of
the discretized device equations, and inference can use either an
equilibrium or a transient trajectory. We focus on passive circuit models
with physically admissible component laws and explicit excitation and
readout. Building on the static equilibrium mechanism
of our recent study, we bring it together with nonlinear RC diffusion
and nonlinear waves in a common circuit formulation. This formulation gives a common
accuracy-energy design objective and operating-cycle accounting across
the three mechanisms, including the input and readout interfaces.

Khoram et al.~\cite{khoram2019nanophotonic} used adjoints of
nonlinear Maxwell equations to optimize the linear-material distribution
of a continuous nanophotonic medium for image inference in simulation,
while keeping the nonlinear regions fixed.

Photonic implementations provide experimental demonstrations using coherent
optical circuits~\cite{shen2017deep}, including integrated chips that combine
linear transformations with nonlinear
processing~\cite{bandyopadhyay2024singlechip}. More recently, Unconventional
AI's Un-0 technical release~\cite{unconventional2026un0} explored image
generation through simulated coupled oscillators. These examples suggest
that different physical mechanisms can serve different computational tasks.
As in neural network design, we choose the architecture to suit the task;
here the choices include the operating physics, geometry, and input and
output interfaces.

\textbf{Problem formulation.} To make the design problem precise, let
\begin{equation}
  S_\theta(u,E(x))=0
  \label{eq:sim}
\end{equation}
describe a physical device driven by encoded input data $E(x)$, with internal
states $u$ and design parameters $\theta$. We consider differential-equation
models, including any algebraic constraints and initial or boundary conditions
in $S_\theta$. The encoder $E$ converts the data $x$ into physical inputs,
such as applied currents, sources, or prescribed boundary values. The states
$u$ may vary across the device and in time. An observation operator $Q$
selects measurements at specified locations or times, or integrates the
response over a time interval. A decoder $\mathcal D$ then converts these
observations into the task output:
$q=Q(u)$ and $y=\mathcal D(q)$.

If the model determines a well-defined response $u_\theta(x)$, the device and
its interfaces define the input--output map
\begin{equation}
  \begin{aligned}
    F_\theta(x) &= \mathcal D\bigl(Q(u_\theta(x))\bigr),\\
    S_\theta\bigl(u_\theta(x),E(x)\bigr) &= 0.
  \end{aligned}
  \label{eq:computational-map}
\end{equation}
Learning consists of choosing physically realizable parameters $\theta$ so
that $F_\theta$ approximates the desired computational map. The physical
model specifies the available interactions and nonlinear responses, while
the parameters determine the particular transformation.

We develop this approach using electrical circuits and structured nonlinear
materials. For circuits, the design variables include connectivity, component
values, and nonlinear current--voltage relations. For materials, they include
geometry and spatially varying constitutive properties. We formulate the
design as an optimization problem constrained by the device equations.
Numerical simulation predicts the response to encoded inputs, and derivatives
of that response are used to adjust the design parameters. This stage can be
computationally demanding. Once built, the device produces its response
through physical operation, and measurement and decoding provide the output.
Our aim is to connect a computational task to a physical design, with energy
expenditure guiding the choice of system and its parameters.

This paper is a computational study: we evaluate the proposed circuits in
simulation. Fabrication and hardware validation are beyond its scope.

\section{Differential equations from electrical circuits}
\label{sec:circuits}

In this section, we derive circuit models for static, diffusive, and wave
computation. The circuit specifies both the governing equations and their
physical realization: connectivity determines which states interact, component
laws determine the interactions, and ports specify how signals enter and leave.
We begin with resistive networks and then add capacitive and inductive storage.
Throughout, $u$ denotes an internal, or hidden, state coordinate. The device
equations determine this coordinate, and observations of the resulting state
provide the output.

\subsection{Why electrical circuits?}
\label{sec:why-circuits}

We choose circuits for three reasons. First, they allow us to realize several
types of physics within a common construction. Resistors define equilibrium
problems. Capacitors store charge and introduce first-order dynamics, while
inductors provide magnetic storage and permit oscillation and wave propagation.
Kirchhoff's laws give a common way to derive these models from the circuit
graph~\cite{dorfler2018electrical}.

Second, circuits provide a practical route from a model to a device. Their
components can be fabricated, connected to sources, and measured using
established electrical interfaces. The response can be fast, with characteristic
scales such as $RC$ for relaxation and $\sqrt{LC}$ for oscillation. Component
values therefore give us control over the time scales of the computation.
A finite circuit also allows a proposed design to be tested before considering
a more elaborate material realization.

Third, spatially local circuits connect naturally to continuous media.
Differences between neighboring node states approximate spatial derivatives.
With suitable geometry and component scaling, resistive arrays approximate
conduction equations, resistor--capacitor arrays approximate diffusion equations,
and inductor--capacitor arrays approximate wave equations. The continuum
interpretation of transmission lines is a familiar
example~\cite{vool2017circuits}. These limits provide a connection between a
discrete circuit design and a structured nonlinear material.

\subsection{Resistors and diodes: static computation}
\label{sec:static-circuits}

Consider a circuit with $n$ unknown node voltages and $m$ branches. We fix a
reference node at zero potential and assume that every other node is connected
to it by a path. Let $u\in\mathbb R^n$ collect the unknown voltages and let
$D\in\mathbb R^{m\times n}$ be the reduced incidence matrix. For a branch $e$
oriented from node $i$ to node $j$, its row is chosen so that
\begin{equation}
  v_e=(Du)_e=u_i-u_j.
  \label{eq:branch-voltage}
\end{equation}
The reference voltage is omitted from $u$. Thus a branch to ground contributes
a row with a single nonzero entry. Let $i\in\mathbb R^m$ collect the branch
currents, with each current positive in its branch orientation. Then $D^Ti$
is the vector of net currents leaving the nodes through the branches.
The pair $(u,i)$ gives a mixed, or flux, description: $u$ is the nodal
potential and $i$ is the current carried by the connections (see
\cref{fig:incidence-example}).

\begin{figure}[htbp]
  \centering
  \begin{tikzpicture}[font=\small, line width=0.6pt]
    \coordinate (a) at (0,2);
    \coordinate (b) at (3,2);
    \coordinate (g) at (1.5,0);
    \draw[->,shorten <=3pt,shorten >=3pt]
      (a) -- node[above] {$e_1,\ i_1$} (b);
    \draw[->,shorten <=3pt,shorten >=3pt]
      (a) -- node[left] {$e_2,\ i_2$} (g);
    \draw[->,shorten <=3pt,shorten >=3pt]
      (b) -- node[right] {$e_3,\ i_3$} (g);
    \fill (a) circle (1.7pt) node[above left] {$u_1$};
    \fill (b) circle (1.7pt) node[above right] {$u_2$};
    \fill (g) circle (1.7pt) node[left=5pt] {$u_0=0$};
    \draw (g) -- (1.5,-0.25);
    \draw (1.2,-0.25) -- (1.8,-0.25);
    \draw (1.3,-0.35) -- (1.7,-0.35);
    \draw (1.4,-0.45) -- (1.6,-0.45);
    \node[anchor=west] at (4.1,1) {$
      \begin{aligned}
        D&=\begin{bmatrix}
          1&-1\\
          1&0\\
          0&1
        \end{bmatrix},\\[0.5em]
        D^Ti&=\begin{bmatrix}
          i_1+i_2\\
          -i_1+i_3
        \end{bmatrix}.
      \end{aligned}$};
  \end{tikzpicture}
  \caption{An oriented circuit graph with two unknown voltages and three
  branches. The rows of $D$ correspond to $e_1,e_2,e_3$ and its columns
  to $u_1,u_2$; the grounded node is omitted. Arrows define positive
  branch currents, and $D^Ti$ gives the net current leaving each unknown node.}
  \label{fig:incidence-example}
\end{figure}

Let $x\in\mathbb R^p$ be the input data. We encode $x$ as injected currents
$f=Bx$, where $B\in\mathbb R^{n\times p}$ specifies the input ports and their
scalings. A column of $B$ may, for example, inject a current at one node and
withdraw it at another, or use the reference node as the return. Thus $B$
describes the source connections and the scaling of the applied currents.

For linear resistors, Ohm's law and Kirchhoff's current law give
\begin{equation}
  \begin{aligned}
    D^Ti&=Bx,\\
    i&=SDu,
  \end{aligned}
  \label{eq:ohm-network}
\end{equation}
where $S=\operatorname{diag}(s_1,\ldots,s_m)$ contains positive branch
conductances. The first equation balances currents at the nodes; the second
relates each current to its voltage drop. Eliminating $i$ gives
\begin{equation}
  D^TSDu=Bx.
  \label{eq:linear-static}
\end{equation}
Because the graph is connected to the reference node, $D$ has full column
rank and $K=D^TSD$ is positive definite. The hidden state is therefore
$u(x)=K^{-1}Bx$. A fixed linear observation and decoder give a linear
input--output map. Nonlinear branch elements allow us to obtain nonlinear
maps.

\begin{figure}[t]
  \centering
\IfFileExists{figures/tanh_diode_branch.png}{%
  \includegraphics[width=\linewidth]{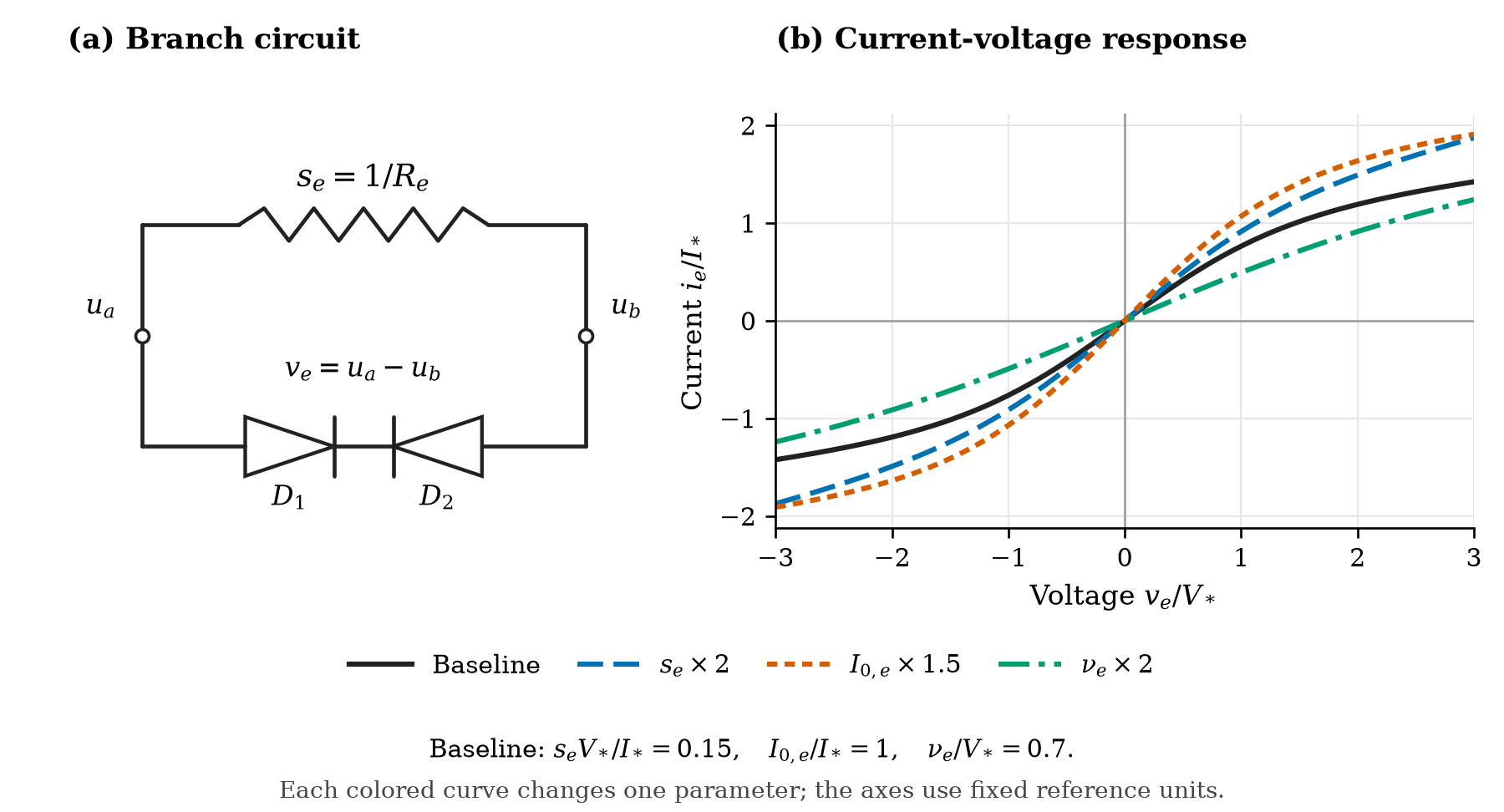}%
}{%
  \includegraphics[width=\linewidth]{figures/tanh_diode_branch.pdf}%
}
  \caption{A resistor in parallel with two matched, oppositely oriented
  series diodes and the resulting current--voltage responses. The axes use
  fixed reference units $V_*$ and $I_*$. Each colored curve changes one
  parameter from the baseline shown below the plots.}
  \label{fig:resistor-diode-tanh}
\end{figure}

One simple choice is a resistor in parallel with two matched diodes connected
in series and oriented in opposite directions, as illustrated in
\cref{fig:resistor-diode-tanh}. Using the quasistatic Shockley law for each diode gives the total branch
current
\begin{equation}
  i_e(v_e)=s_e v_e+
  I_{0,e}\tanh\!\left(\frac{v_e}{2\nu_e}\right),
  \qquad s_e>0,\quad I_{0,e}>0,\quad \nu_e>0.
  \label{eq:resistor-diode}
\end{equation}
Here $\nu_e$ includes the thermal voltage and diode ideality factor, and
$I_{0,e}$ is the saturation current of each diode. The diode pair supplies an odd, saturating nonlinear response, while the
parallel resistor gives a positive linear conductance. The total branch law
is therefore odd and strictly increasing. We assume matched diodes and an
operating range in which reverse breakdown and junction capacitance can be
neglected. The selected devices determine the branch parameters.

More generally, write the branch law as
\begin{equation}
  i=\Phi(SDu),\qquad
  \Phi(z)=\bigl(\phi_e(z_e)\bigr)_{e=1}^m,
  \label{eq:nonlinear-branch}
\end{equation}
where each $\phi_e$ is continuous and nondecreasing, with $\phi_e(0)=0$.
For the branch law in \eqref{eq:resistor-diode},
$\phi_e(z)=z+I_{0,e}\tanh(z/(2s_e\nu_e))$.
The functions act separately on the branches; the graph couples the hidden
states. The nonlinear static model is therefore
\begin{equation}
    D^Ti=Bx,\quad
    i=\Phi(SDu).
  \label{eq:nonlinear-static}
\end{equation}
Eliminating the current recovers $D^T\Phi(SDu)=Bx$.
We use the abbreviation $A(u)=D^T\Phi(SDu)$ when working with the potential
alone. The branch law determines the current from $u$. Retaining $i$ explicitly
separates current balance from the constitutive relations.
The orientation of a diode is part of the physical design and is included in
the branch orientation and constitutive law.

Because the branch laws are monotone, the static equations are the optimality
conditions of a convex energy. Positive parallel conductances make this energy
strictly convex, so the equilibrium exists and is unique for every input
current $Bx$. The circuit therefore defines its hidden state implicitly,
connecting it to deep equilibrium and other implicit
networks~\cite{bai2019deep}, as well as to equilibrium-propagation
models~\cite{scellier2017equilibrium,kendall2020training}. In hardware, the
state is obtained directly through electrical relaxation.

In earlier work
we proved that a class of
current-driven passive resistive networks with grounded threshold elements,
a differential current readout, and a constant offset can uniformly approximate
any continuous map on a compact subset of the nonnegative orthant.

In simulation, the nonlinear equilibrium can be computed with Newton's
method. At each Newton iteration, the linearized Kirchhoff equations require
the solution of a linear system whose dimension is the number $n$ of unknown
node voltages. A dense factorization therefore costs $O(n^3)$ operations and
$O(n^2)$ storage, although sparse circuit graphs can substantially reduce this
cost. This numerical solve is required during digital training and simulation;
the physical device reaches the equilibrium through its own dynamics.

\subsection{Capacitors: diffusion and the heat equation}
\label{sec:rc-circuits}

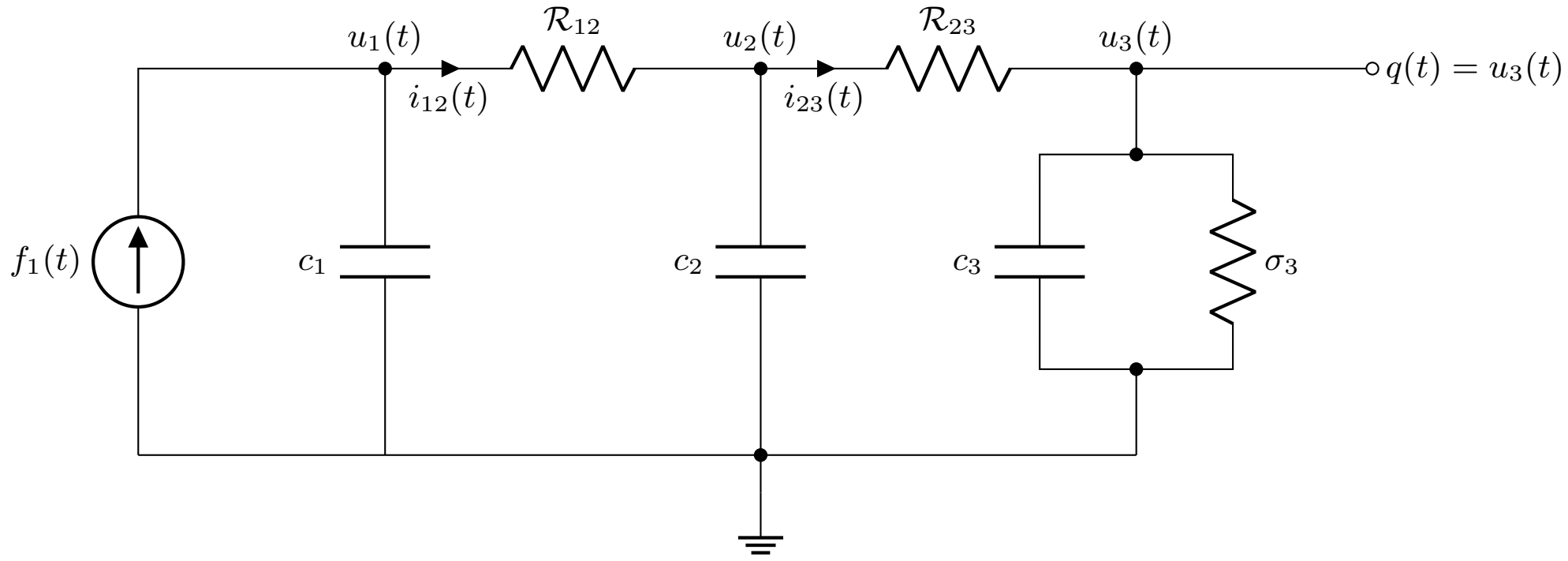
\begin{figure}[htbp]
  \centering
  \begin{circuitikz}[american, scale=0.9, transform shape, font=\small]
    \draw (0,0) to[I,l=$f_1(t)$] (0,3.6) -- (2.3,3.6);

    \draw (2.3,3.6) node[circ] {} node[above] {$u_1(t)$}
      to[R,l=$\mathcal R_{12}$,i>_=$i_{12}(t)$] (5.8,3.6)
      node[circ] {} node[above] {$u_2(t)$}
      to[R,l=$\mathcal R_{23}$,i>_=$i_{23}(t)$] (9.3,3.6)
      node[circ] {} node[above] {$u_3(t)$}
      -- (11.5,3.6) node[ocirc] {}
      node[right] {$q(t)=u_3(t)$};

    \draw (2.3,3.6) -- (2.3,2.8)
      to[C,l_=$c_1$] (2.3,0.8) -- (2.3,0);

    \draw (5.8,3.6) -- (5.8,2.8)
      to[C,l_=$c_2$] (5.8,0.8) -- (5.8,0);

    \draw (9.3,3.6) -- (9.3,2.8) node[circ] {};

    \draw (9.3,2.8) -- (8.4,2.8)
      to[C,l_=$c_3$] (8.4,0.8) -- (9.3,0.8);

    \draw (9.3,2.8) -- (10.2,2.8)
      to[R,l=$\sigma_3$] (10.2,0.8) -- (9.3,0.8);

    \draw (9.3,0.8) node[circ] {} -- (9.3,0);

    \draw (0,0) -- (9.3,0);
    \draw (5.8,0) node[circ] {}
      -- (5.8,-0.35) node[ground] {};
  \end{circuitikz}

  \caption{An RC computing circuit. The resistor symbols
  $\mathcal R_{ab}$ represent nonlinear resistive branches, for example
  those in \cref{fig:resistor-diode-tanh}. Each node has a shunt
  capacitor $c_a$. At node 3, a single lead splits into the parallel
  capacitor $c_3$ and conductance $\sigma_3$, then rejoins before
  reaching ground. The source injects $f_1(t)$, and the output
  measures $u_3(t)$.}
  \label{fig:rc-circuit}
\end{figure}

We next connect a capacitor $c_a>0$ from each unknown node to ground, retaining
the nonlinear resistive branches. The charge at node $a$ is $c_a u_a$, so the
current entering its capacitor is $c_a\dot u_a$. With
$C=\operatorname{diag}(c_a)$, Kirchhoff's law and the resistive branch
law give
\begin{equation}
  \begin{aligned}
    C\dot u+D^Ti&=B(t)x(t),\\
    i&=\Phi(SDu),
  \end{aligned}
  \qquad u(0)=u_{\mathrm{init}}.
  \label{eq:nonlinear-heat}
\end{equation}
Here $D$ includes all resistive branches, including conductances to ground,
and $i(t)$ collects their currents. A shunt conductance $\sigma_a$ draws the
current $\sigma_a u_a$, while $C\dot u$ gives the capacitor currents.
The source current is divided between
charging these capacitors and flowing through the network. Eliminating $i$
gives $C\dot u+D^T\Phi(SDu)=B(t)x(t)$.
The initial state must be specified as part of the device operation. It may be
a fixed reset state or an additional encoding of the input. The matrix $B(t)$
allows sources to be applied at different ports or at different times.
\Cref{fig:rc-circuit} shows a fragment of this construction.

The time-dependent drive $B(t)x$ generates an entire trajectory $u(t)$ rather
than a single equilibrium state. We record this history through
\begin{equation}
  q(t)=Q(t)u(t),
  \label{eq:rc-history-readout}
\end{equation}
where $Q(t)$ selects or combines the voltages of interest. The samples of
$q(t)$ at several times can be concatenated and passed to a learned decoder.
Time therefore provides additional features: different observation times
capture different stages of the nonlinear diffusion, while the input map
$B(t)$ controls how the data excite these responses. The model can learn from
the full transient rather than only from the final, smoothed state.

For numerical simulation, we use implicit Euler. With time step $\Delta t$,
each update satisfies
\begin{equation}
  C\frac{u^{k+1}-u^k}{\Delta t}
  +D^T\Phi(SDu^{k+1})=B(t_{k+1})x.
  \label{eq:rc-implicit-euler}
\end{equation}
This has the same nonlinear structure as the static circuit equation, with the
additional diagonal term $C/\Delta t$. Consequently, each time step can be
solved with the same Newton machinery as the static problem. The RC dynamics
are dissipative and vary smoothly, so a modest number of time steps is
sufficient to resolve the transient features used by the readout.

The capacitor energy and power balance are
\begin{equation}
  H_C(u)=\frac12u^TCu,\qquad
  \frac{dH_C}{dt}=u^TB(t)x-(Du)^Ti.
  \label{eq:rc-power}
\end{equation}
Source work supplies both stored energy and resistive losses.

\subsection{Inductors: nonlinear second-order dynamics}
\label{sec:rlc-circuits}

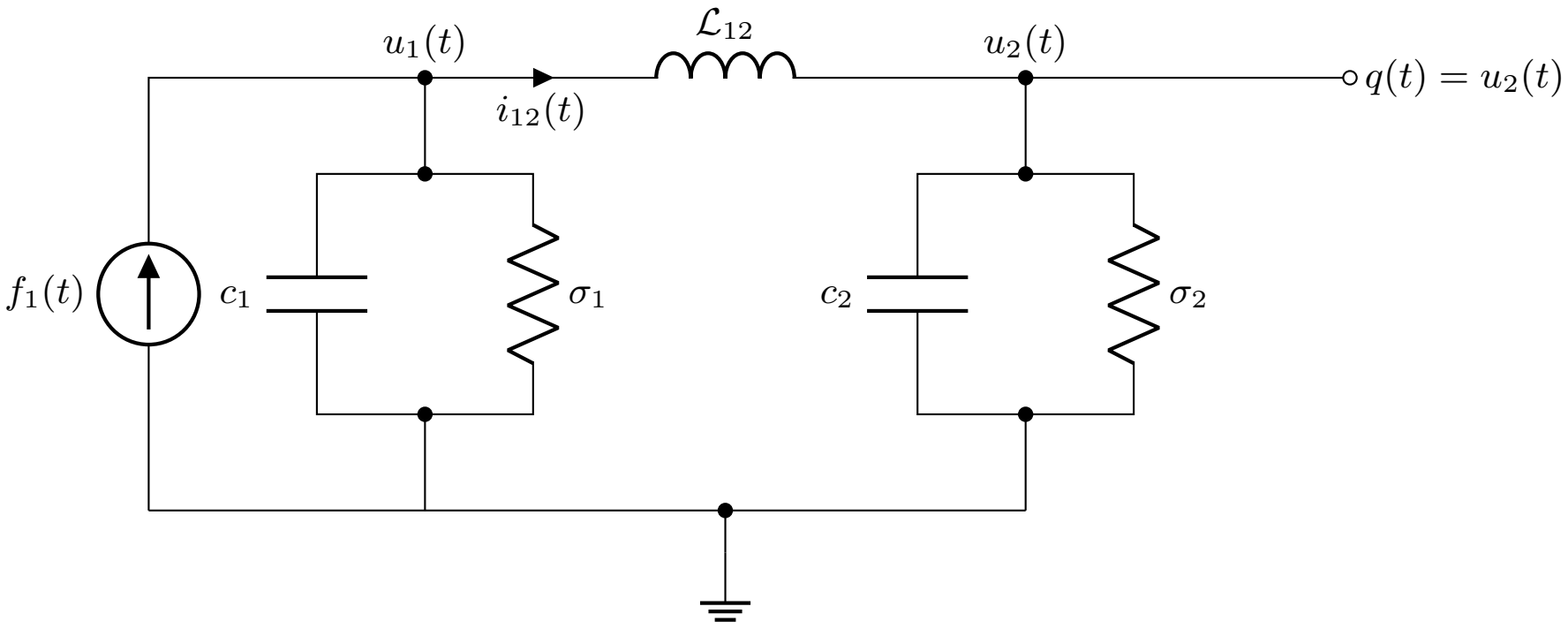
\begin{figure}[htbp]
  \centering
  \begin{circuitikz}[american, scale=0.9, transform shape, font=\small]
    \draw (0,0) to[I,l=$f_1(t)$] (0,3.6) -- (2.3,3.6);
    \draw (2.3,3.6) node[circ] {} node[above] {$u_1(t)$}
      to[L,l=$\mathcal L_{12}$,i>_=$i_{12}(t)$] (7.3,3.6)
      node[circ] {} node[above] {$u_2(t)$}
      -- (10,3.6) node[ocirc] {} node[right] {$q(t)=u_2(t)$};
    \draw (2.3,3.6) -- (2.3,2.8) node[circ] {};
    \draw (2.3,2.8) -- (1.4,2.8)
      to[C,l_=$c_1$] (1.4,0.8) -- (2.3,0.8);
    \draw (2.3,2.8) -- (3.2,2.8)
      to[R,l=$\sigma_1$] (3.2,0.8) -- (2.3,0.8);
    \draw (2.3,0.8) node[circ] {} -- (2.3,0);
    \draw (7.3,3.6) -- (7.3,2.8) node[circ] {};
    \draw (7.3,2.8) -- (6.4,2.8)
      to[C,l_=$c_2$] (6.4,0.8) -- (7.3,0.8);
    \draw (7.3,2.8) -- (8.2,2.8)
      to[R,l=$\sigma_2$] (8.2,0.8) -- (7.3,0.8);
    \draw (7.3,0.8) node[circ] {} -- (7.3,0);
    \draw (0,0) -- (7.3,0);
    \draw (4.8,0) node[circ] {} -- (4.8,-0.35) node[ground] {};
  \end{circuitikz}
  \caption{Nonlinear inductive coupling and parallel shunts. A single lead
  from each node splits into a capacitor $c_a$ and a resistive branch
  labelled by its conductance $\sigma_a$, then rejoins before reaching
  ground. The source injects $f_1(t)$ and the output measures $u_2(t)$.}
  \label{fig:wave-circuit}
\end{figure}

We place nonlinear inductors on the graph edges and connect a linear
capacitor and conductance in parallel from each node to ground, as shown
in \cref{fig:wave-circuit}. The inductors couple neighboring states; the
shunts provide local storage and dissipation. We retain $u$ for the node
voltages and $i$ for the oriented edge currents. The incidence matrix $D$
contains the inductive edges, while the shunt parameters are collected in
\begin{equation}
  C=\operatorname{diag}(c_a),\qquad
  S=\operatorname{diag}(\sigma_a),\qquad c_a,\sigma_a>0.
  \label{eq:shunt-matrices}
\end{equation}
The corresponding shunt currents are $C\dot u$ and $Su$.

An inductor relates its current to its flux linkage $\lambda_e$, whose
time derivative is the voltage drop $(Du)_e$~\cite{vool2017circuits}.
We write $i_e=\phi_e(\lambda_e/\ell_e)$, where $\ell_e>0$ is a reference
inductance and $\phi_e$ is smooth and strictly increasing, with
$\phi_e(0)=0$. We neglect hysteresis. The choice $\phi_e(z)=z$ recovers a linear inductor.
A tanh model of magnetic saturation~\cite{ricci2021linear} can be combined
with a linear flux contribution to give, for example,
\begin{equation}
  \frac{\lambda_e}{\ell_e}
    =i_e+a_e\tanh(i_e/b_e),\qquad a_e,b_e>0.
  \label{eq:tanh-inductor}
\end{equation}
Here $a_e$ and $b_e$ are current scales. The inverse of the increasing map
$i\mapsto i+a_e\tanh(i/b_e)$ defines $\phi_e$. Thus the saturation law
is expressed as flux versus current, and its inverse supplies the restoring
current.

With $L=\operatorname{diag}(\ell_e)$ and injected currents
$f(t)=B(t)x(t)$, the circuit equations are
\begin{equation}
  \begin{aligned}
    C\dot u+Su+D^Ti&=f(t),\\
    \dot\lambda&=Du,\\
    i&=\Phi(L^{-1}\lambda),
  \end{aligned}
  \qquad \lambda(0)=0,\quad u(0)=u_{\mathrm{init}}.
  \label{eq:nonlinear-wave}
\end{equation}
The first equation balances the capacitor, conductance, and inductive
currents at each node. We assume zero initial flux, corresponding to
resetting the inductors before applying an input.

Define the node flux by
\begin{equation}
  \psi(t)=\int_0^t u(\tau)\,d\tau,\qquad
  \dot\psi=u,\qquad \lambda=D\psi.
  \label{eq:node-flux}
\end{equation}
Substitution into current balance gives
\begin{equation}
  \begin{aligned}
    C\ddot\psi+S\dot\psi+D^T\Phi(L^{-1}D\psi)&=B(t)x(t),\\
    \psi(0)=0,\qquad \dot\psi(0)&=u_{\mathrm{init}}.
  \end{aligned}
  \label{eq:nonlinear-wave-restoring}
\end{equation}
The capacitances and conductances give diagonal mass and damping matrices.
The graph enters only through the restoring term: $D\psi$ gives the flux
differences, $\Phi$ gives the inductive currents, and $D^T$ assembles them
at the nodes. For a local array with suitable component scaling, this is
a discretization of a nonlinear damped wave equation.

We discretize \eqref{eq:nonlinear-wave-restoring} with a damped leapfrog
scheme. Centering both the acceleration and damping terms gives
\begin{equation}
  C\frac{\psi^{k+1}-2\psi^k+\psi^{k-1}}{\Delta t^2}
  +S\frac{\psi^{k+1}-\psi^{k-1}}{2\Delta t}
  +D^T\Phi(L^{-1}D\psi^k)=B(t_k)x(t_k).
  \label{eq:nonlinear-wave-leapfrog}
\end{equation}
Because $C$ and $S$ are diagonal, computing $\psi^{k+1}$ requires only the
evaluation of the nonlinear branch currents and a diagonal solve. The method
therefore advances the physical state without forming or factorizing a dense
matrix.

The time step is limited by a Courant-Friedrichs-Lewy (CFL) condition. The
linearized stiffness at state $\psi$ is
\begin{equation}
  K(\psi)=D^T\operatorname{diag}\!\left(
  \frac{1}{\ell_e}\phi'_e\bigl((L^{-1}D\psi)_e\bigr)
  \right)D.
  \label{eq:nonlinear-wave-stiffness}
\end{equation}
For the undamped leapfrog scheme, stability requires
\begin{equation}
  \Delta t^2\lambda_{\max}\!\left(C^{-1}K(\psi)\right)\leq 4.
  \label{eq:nonlinear-wave-cfl}
\end{equation}

For ideal inductors, let
$\Psi_e(z)=\int_0^z\phi_e(\zeta)\,d\zeta$. The stored energy satisfies
\begin{equation}
  \begin{aligned}
    H(u,\lambda)&=\frac12u^TCu+
      \sum_e\ell_e\Psi_e(\lambda_e/\ell_e),\\
    \frac{dH}{dt}&=u^TB(t)x(t)-u^TSu.
  \end{aligned}
  \label{eq:nonlinear-wave-energy}
\end{equation}
This ideal model includes shunt dissipation but excludes winding and magnetic
core losses, which would also alter the dynamics.

\section{Learning and physical design}
\label{sec:learning-design}

Given training pairs $\{(x_j,y_j)\}_{j=1}^N$, we seek a device whose response
to $x_j$ predicts $y_j$. The design includes the circuit, the encoding of
the input, and the decoding of the measured response. These choices must
be made together: the same circuit can perform different computations
when driven or observed through different ports.

We return to the notation of the introduction and write $u_\theta(x)$ for
the complete hidden state. This includes the node voltages and any currents
or fluxes needed to describe the device; for a dynamic model, it includes
their time history. To simplify notation, we absorb the encoder
$E_\theta$ and the source map $B_\theta$ into the simulation operator:
\begin{equation}
  S_\theta\bigl(u_\theta(x),x\bigr)=0,\qquad
  F_\theta(x)=\mathcal D_\theta\bigl(Q_\theta(u_\theta(x))\bigr).
  \label{eq:design-forward-map}
\end{equation}
The operator $S_\theta$ includes the circuit
equations and their initial or boundary conditions. The observation
$Q_\theta$ specifies what is measured, and $\mathcal D_\theta$ maps those
measurements to the prediction. Fixed interfaces are obtained by holding
the corresponding parameters constant.

\subsection{Architecture and admissible parameters}
\label{sec:design-parameters}

Before tuning component values, we must choose an architecture. This
includes the type of circuit, the number and arrangement of nodes, the
available nonlinear elements, and the input and output ports. For dynamic
devices, the initial state, operating time, and measurement times are
also part of the specification. A static circuit computes through an
equilibrium, whereas an RC or inductive circuit can use its transient
response. These choices determine the family of maps available for learning.

For a fixed architecture, $\theta$ collects the component values,
constitutive parameters, and any trainable parameters in $E_\theta$,
$B_\theta$, $Q_\theta$, and $\mathcal D_\theta$. The admissible set
$\Theta_{\mathrm{phys}}$ describes the realizable designs. Capacitances
and inductances must be positive, passive conductances must be
nonnegative, and component values lie within attainable ranges.
Nonlinear laws must come from available components or materials. For
the passive models above, we preserve their monotonicity and storage
properties while adjusting their parameters. Wiring, fabrication, and
access to ports impose further restrictions. The selected component
models must also remain valid over the voltages and currents reached
during operation.
The normalized equations need not determine unique resistance and capacitance
scales; we optimize these values within their physically admissible ranges
in $\Theta_{\mathrm{phys}}$.

The architecture itself can be designed. In particular, choosing the
connections changes the incidence matrix $D$. Its entries encode a
graph and cannot be varied as arbitrary real numbers. One approach is
to search over a family of graphs and optimize the continuous parameters
for each candidate. Another is to begin with a larger admissible graph,
learn nonnegative edge weights, and remove weak connections by introducing penalty in the loss function. The resulting
circuit must then be checked and, if needed, optimized again. 

\subsection{An objective for accuracy and energy}
\label{sec:design-objective}

Let $\ell_j(\theta,u)=
\ell(\mathcal D_\theta(Q_\theta(u)),y_j)$ measure prediction error, and
let $\mathcal E_j(\theta,u)$ be the energy required to process $x_j$.
We formulate the design problem as
\begin{equation}
  \begin{aligned}
\min_{\substack{\theta\in\Theta_{\mathrm{phys}}\\u_1,\ldots,u_N}}
    \quad &\frac1N\sum_{j=1}^N
      \left[\ell_j(\theta,u_j)+\beta\mathcal E_j(\theta,u_j)\right]
      +\alpha\mathcal R(\theta),\\
    \text{subject to}\quad&S_\theta(u_j,x_j)=0,
      \qquad j=1,\ldots,N.
  \end{aligned}
  \label{eq:physical-design-problem}
\end{equation}
Here $\beta\geq0$ controls the tradeoff between prediction error and
energy. The
regularizer $\mathcal R$ can favor simpler designs by applying sparsity, and $\alpha\geq0$ sets its weight.
Including energy in the objective allows the design to sacrifice a
small improvement in accuracy when obtaining it would require a large
increase in energy. Related tradeoffs have been studied in physical
learning circuits~\cite{stern2024power}.

Energy must refer to a specified operating cycle. We include input
encoding, excitation, sensing, decoding, and restoration to the initial
state. The balances in \eqref{eq:rc-power} and \eqref{eq:nonlinear-wave-energy}
provide the circuit contribution. For example, over an interval $[0,T]$
the resistive losses are
\begin{equation}
  \mathcal E_{\mathrm{RC,diss}}=
    \int_0^T(Du)^Ti\,dt,\qquad
  \mathcal E_{\mathrm{L,diss}}=
    \int_0^T u^TSu\,dt
  \label{eq:design-dissipation}
\end{equation}
for the RC and inductive models, respectively. These losses and the
change in stored energy determine the work supplied at the circuit
ports; they are not additional costs to be added to that same work.
The supply and reset models determine whether stored energy is
recovered or dissipated. A static equilibrium alone does not specify
energy per evaluation: settling, holding, and reset times must also
be prescribed. The interfaces must be included in this accounting,
since a complicated encoder or decoder can consume more energy than
the circuit. Signal levels, sensing accuracy, and operating time must
likewise be specified when comparing designs.

The design problem in \eqref{eq:physical-design-problem} is an optimization problem
constrained by the device equations. It has the same structure as
PDE-constrained optimization and nonlinear inverse
problems~\cite{hinze2009optimization,haber2000optimization}.
The states are determined by a simulation, while the parameters are
shared across all training examples. Eliminating the states gives the
reduced objective
\begin{equation}
  J(\theta)=\frac1N\sum_{j=1}^N
    \left[\ell_j(\theta,u_\theta(x_j))+
      \beta\mathcal E_j(\theta,u_\theta(x_j))\right]
      +\alpha\mathcal R(\theta).
  \label{eq:reduced-design-objective}
\end{equation}
Evaluating $J$ requires solving the device equations. Differentiating
it requires accounting for the dependence of the hidden states on the
design.

\subsection{Lagrange multipliers and the adjoint}
\label{sec:design-adjoint}

We derive the gradient after discretizing any time or spatial variables.
Thus $u_j$ denotes the complete discrete state, and $S_\theta(u_j,x_j)$
is the corresponding residual. We assume differentiable constitutive
laws and a locally well-defined response, with a nonsingular state
Jacobian. Introduce a multiplier $p_j$ for each simulation and form
the Lagrangian
\begin{equation}
  \mathcal L(\theta,\{u_j\},\{p_j\})=
    \alpha\mathcal R(\theta)+\frac1N\sum_{j=1}^N
    \left[\ell_j(\theta,u_j)+\beta\mathcal E_j(\theta,u_j)
      +p_j^TS_\theta(u_j,x_j)\right].
  \label{eq:design-lagrangian}
\end{equation}
Stationarity with respect to $p_j$ recovers the device equations.
Stationarity with respect to $u_j$ gives the adjoint equations
\begin{equation}
  \bigl(\partial_u S_\theta(u_j,x_j)\bigr)^T p_j
    =-\nabla_u\ell_j(\theta,u_j)
      -\beta\nabla_u\mathcal E_j(\theta,u_j).
  \label{eq:design-adjoint}
\end{equation}
Once the states and adjoints have been computed, the reduced gradient is
\begin{equation}
  \nabla J(\theta)=\alpha\nabla\mathcal R(\theta)+\frac1N\sum_{j=1}^N
    \left[\nabla_\theta\ell_j+
      \beta\nabla_\theta\mathcal E_j+
      \bigl(\partial_\theta S_\theta(u_j,x_j)\bigr)^T p_j\right].
  \label{eq:design-gradient}
\end{equation}
The partial derivatives in the brackets hold $u_j$ fixed. They include
the direct dependence of the loss and energy on the decoder and other
parameters. The last term accounts for changes in the circuit response,
including changes to the encoder and source map within $S_\theta$.
At an interior stationary point this gradient vanishes; at a component
bound, it satisfies the corresponding one-sided optimality conditions.

The adjoint avoids computing a separate state sensitivity for every
parameter. One forward simulation and one adjoint solve per example
give the gradient of the combined accuracy and energy objective.
For a static circuit, the adjoint uses the transpose of the linearized
equilibrium operator. For a dynamic circuit, it runs backward through
the discretized evolution, with contributions from the readout and
the energy accumulated along the trajectory. Using the adjoint of the
chosen time discretization gives the derivative of the simulation
actually used in the objective. State trajectories must be stored or
recomputed as needed.

\subsection{Solving the design problem}
\label{sec:design-algorithms}

The choice of optimizer depends on the number of training examples,
the cost of simulation, and the accuracy required in the final design.
For large data sets, stochastic gradient descent uses a small batch
of examples to estimate the gradient in \eqref{eq:design-gradient}.
The same batch contributes both prediction and energy derivatives.
Forward and adjoint solves for different examples are independent.
Component bounds can be enforced by projection or by a suitable
parameterization. Stochastic methods reduce the cost per update, but
gradient noise can slow accurate final
optimization~\cite{bottou2018optimization}.

When reliable gradients are affordable, ADAM is routinely used. We have also experimented with limited-memory BFGS that uses
successive parameter and gradient differences to approximate curvature.
It requires no explicit Hessian and uses curvature information to scale
the parameter updates. The bound-constrained variant L-BFGS-B is
particularly useful for component ranges~\cite{byrd1995limited}.

The resulting
design is assessed on inputs withheld from training, using both
prediction error and energy per operating cycle. While simulation and
optimization can be computationally intensive, they only take place during design; the fabricated device evaluates
the learned map through its physical response. Any energy saving depends on
the physical realization and the cost of its interfaces and readout.

\section{Numerical experiments}
\label{sec:experiments}

We consider three experiments: classification with a static circuit,
image denoising with an RC circuit, and language modeling with an
inductive circuit. The experiments test how nonlinear
equilibria, diffusive relaxation, and oscillatory memory can be
used for different tasks. In each case, we  examine the relation
between prediction quality, energy expenditure, and operating time.

\subsection{Experimental setup and energy accounting}
\label{sec:experiment-setup}

We train the physical devices using \eqref{eq:physical-design-problem}. After a little experimentation, we set the energy and design-regularization weights to $\beta=\alpha=0.01$.
The circuit parameterizations enforce passivity, and resistance and capacitance
values are chosen within physically admissible component ranges. Input amplitudes, sensing
precision, and latency budgets are held fixed when comparing
energy weights within an experiment. The encoders and decoders
are affine or simple (one layer) maps, or embedding lookups, so that the majority of the prediction
is performed by the circuit.

Model selection uses validation data only. Numerical resolution is checked
by refinement. The number of
independent trainable parameters includes the interfaces.
The number of physical nodes and edges is also reported.

Energy accounting refers to a complete evaluation cycle,
including input preparation, excitation, sensing, decoding, and
reset. Circuit contributions follow from the balances in
\eqref{eq:rc-power} and \eqref{eq:nonlinear-wave-energy}, together with the driver
and interface models. Where physical scales remain unspecified, we report
electronics subtotals and identify the uncalibrated medium contribution
separately.
Predicted device energy, measured digital
baseline energy, and the runtime of the numerical simulator are
identified separately. 

For model-based digital estimates, we use
$e_{\mathrm{MAC}}=4.6\times10^{-12}\,\mathrm J$ per float32 MAC and
$e_{\mathrm r}=5\times10^{-12}\,\mathrm J$ per 32-bit local memory read,
using Horowitz's historical primitive costs as a
reference~\cite{horowitz2014energy}. We assume banked on-chip weights and
local reuse of activations, with no off-chip traffic. Our reference budget is
\begin{equation}
  \widehat{\mathcal E}_{\mathrm{dig}}
  =1.1\left(e_{\mathrm{MAC}}n_{\mathrm{MAC}}
       +e_{\mathrm r}n_{\mathrm r}\right).
  \label{eq:digital-energy-budget}
\end{equation}
The additional 10\% is an explicit allowance for nonlinearities, activation
traffic, and control. For each driven or sensed port we allocate
$e_{\mathrm c}=10^{-11}\,\mathrm J$ for conversion at up to eight bits and
$e_{\mathrm b}=10^{-11}\,\mathrm J$ for buffering and interface control per
evaluation. These are engineering allocations, not measured device costs or
guaranteed bounds; standby leakage is excluded. The total circuit energy is
the electronics budget plus the work drawn to excite, hold, and reset the
medium. 

\subsection{Static bipartite circuit for classification}
\label{sec:experiment-classification}

We use the UCI Letter Recognition dataset~\cite{frey1991letter}, a
benchmark for multiclass classification. It contains 20{,}000 examples of the
26 capital letters of the English alphabet, generated from images of letters
in 20 fonts with random distortions. Each example is represented by 16 integer
attributes describing the size of the letter, the distribution of its pixels,
and its edge counts. We use the first 16{,}000 examples for training and the
remaining 4{,}000 for testing. Thus, the experiment begins with the supplied
feature vector; recovering these features from an image is not part of the
classification task.

\begin{figure}[t]
  \centering
  \resizebox{0.92\linewidth}{!}{%
  \begin{tikzpicture}[font=\small, line width=0.55pt,
      port/.style={draw, rounded corners=3pt, align=center,
                   minimum height=1.0cm, minimum width=1.8cm},
      state/.style={circle, draw=mediumdraw, fill=white, inner sep=0pt,
                    minimum size=5.5pt},
      level/.style={draw=mediumdraw, fill=mediumfill, rounded corners=4pt,
                    minimum width=0.75cm, minimum height=3.0cm}]
    \node[port, draw=inputdraw, fill=inputfill] (data) at (-2.6,0)
      {features\\$x\in\mathbb R^{16}$};
    \node[port, draw=inputdraw, fill=inputfill] (enc) at (0,0)
      {source map\\$B$};

    \foreach \ell/\xx in {0/2.4,1/5.0,2/7.6} {
      \node[level] (lev\ell) at (\xx,0) {};
      \foreach \j/\yy in {1/1.15,2/0.38,3/-0.38,4/-1.15} {
        \node[state] (n\ell\j) at (\xx,\yy) {};
      }
    }
    \foreach \a in {1,2,3,4} {
      \foreach \b in {1,2,3,4} {
        \draw[mediumdraw!45] (n0\a) -- (n1\b);
        \draw[mediumdraw!45] (n1\a) -- (n2\b);
      }
    }
    \foreach \ell in {0,1,2} {
      \foreach \j in {1,2,3,4} {
        \node[state] at (n\ell\j) {};
      }
    }

    \node[port, draw=readdraw, fill=readfill] (read) at (10.0,0)
      {logit\\interface};
    \node[port, draw=outdraw, fill=outfill] (logits) at (12.7,0)
      {logits\\$z\in\mathbb R^{26}$};
    \draw[->, thick, inputdraw] (data) -- (enc);
    \draw[->, thick, inputdraw] (enc) -- node[above] {$Bx$} (lev0);
    \draw[->, thick, readdraw] (lev2) -- (read);
    \draw[->, thick, readdraw] (read) -- (logits);
    \node[above, text depth=0.25ex] at (2.4,1.55) {left level};
    \node[above, text depth=0.25ex] at (5.0,1.55) {hidden level};
    \node[above, text depth=0.25ex] at (7.6,1.55) {right level};
    \node[below, align=center] at (5.0,-1.6)
      {monotone resistor--diode branches\\between consecutive levels};
  \end{tikzpicture}%
  }
  \caption{Static bipartite classifier. The 16-dimensional feature vector is
  encoded as currents at the left boundary \textcolor{inputdraw}{(blue)}. The nonlinear passive
  network \textcolor{mediumdraw}{(green)}
  relaxes to equilibrium, and voltages at the right boundary are converted
  by the logit interface \textcolor{readdraw}{(orange)} into 26 class logits \textcolor{outdraw}{(purple)}.
  All-to-all interlevel lines depict the dense model;
  the sparse model retains one in twelve candidate branches. Only four nodes
  per level are drawn.}
  \label{fig:static-bipartite-classifier}
\end{figure}
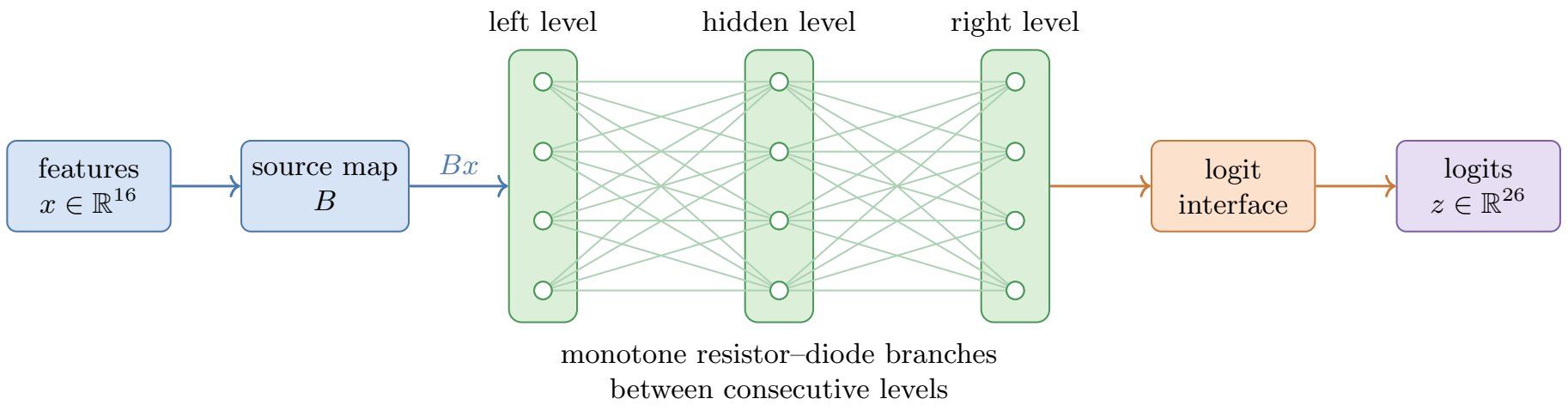

The classifier is the static bipartite network shown in
\cref{fig:static-bipartite-classifier}. Its nodes are divided into three
levels, and physical branches join only consecutive levels. In the dense
model, every node in one level is connected to every node in the next. In the
sparse model, the same layered construction is used but only one in twelve
possible interlevel branches is retained. We selected these branches at random
before training and kept the connectivity fixed throughout training.
The corresponding level widths are
$(96,64,96)$ for the dense network and $(96,576,96)$ for the sparse network;
the larger middle level in the sparse model gives the two models comparable
numbers of trainable parameters.

The feature vector $x\in\mathbb R^{16}$ enters through the nodes at one end of
the network as the source current $Bx$. The circuit then relaxes to the
nonlinear equilibrium in \eqref{eq:nonlinear-static}. Each interlevel branch is
a monotone resistor--diode element, and positive shunt conductances make the
equilibrium well posed. At the opposite end, the observation $q=Q(u)$ is
decoded into a vector of 26 logits, $z=\mathcal D(q)$. Thus the data enter from
one side of the bipartite network and the logits are read from the other. The
source map, circuit parameters, and decoder are trained jointly using the
cross-entropy loss.

The baselines are an MLP and a ResNet of comparable size.
\Cref{tab:classification-results} reports prediction accuracy separately from
the implementation-dependent energy and latency budgets in
\cref{tab:classification-energy}.

We use batch size one to represent low-latency inference on individual inputs,
without waiting to accumulate a batch. GPU energies are measurements that
include execution overheads, whereas circuit energies are model-based
budgets computed from the device and interface models, including settling and
reset. The comparison is therefore specific to the measured GPU operating
point and the assumed circuit realization.
We compute each circuit budget
term by term. The digital source map and readout follow
\eqref{eq:digital-energy-budget}, each driven or sensed port adds the
conversion and buffering allocations of \cref{sec:experiment-setup}, and the
medium adds the source work of the simulated settling from reset.  

The sparse circuit matches the MLP and is 0.2 percentage points below the
ResNet, while improving on the dense circuit by 0.4 percentage points.
For a reference implementation with digital affine interfaces, the source
map and readout require $16\times96+96\times26=4{,}032$ MACs per example.
Their separate budgets, including conversion and buffering at 96 ports on
each side, are $1.8\times10^{-8}\,\mathrm J$ for the input interface and
$2.8\times10^{-8}\,\mathrm J$ for the readout. With a provisional circuit
source-work contribution of $10^{-12}$--$10^{-11}\,\mathrm J$, this gives
approximately $4.6\times10^{-8}\,\mathrm J$ per example
for either circuit. This source-work estimate and the retained settling
times assume $10^{-15}\,\mathrm F$ parasitic capacitance per node and remain
conditional on the circuit model. Other input and readout implementations
would change their respective budget terms.

\begin{table}[t]
  \caption{Letter Recognition prediction results. Accuracy is the test mean
  $\pm$ standard deviation over three training seeds. Hardware-dependent
  energy and latency are reported separately in
  \cref{tab:classification-energy}.
  }
  \label{tab:classification-results}
  \centering
  \small
  \renewcommand{\arraystretch}{1.2}
  \begin{tabular*}{\linewidth}{@{\extracolsep{\fill}}lrc@{}}
    \toprule
    Model & Parameters & Accuracy (\%)\,$\uparrow$\\
    \midrule
    MLP & 32,303 & $97.98 \pm 0.18$\\
    ResNet & 32,126 & $98.12 \pm 0.04$\\
    \midrule
    Nonlinear static, dense (ours)
      & 32,570 & $97.56 \pm 0.07$\\
    Nonlinear static, sparse (ours)
      & 28,988 & $97.94 \pm 0.28$\\
    \bottomrule
  \end{tabular*}
\end{table}

\begin{table}[t]
  \caption{Energy per classified example and energy efficiency for
  Letter Recognition. GPU measurements use an NVIDIA RTX 4090 in
  float32 with batch size one. Circuit values are model-based
  estimates that include digital interfaces and medium source work.
  MAC-model estimates use \eqref{eq:digital-energy-budget} and assume
  one weight read per MAC.}
  \label{tab:classification-energy}
  \centering
  \begin{tabular}{lrr}
    \toprule
    Model & Energy (J/example) & Efficiency (examples/J) \\
    \midrule
    MLP 
      & $3.4\times10^{-3}$ & $2.9\times10^{2}$ \\
    ResNet 
      & $1.2\times10^{-2}$ & $8.3\times10^{1}$ \\
    MLP (MAC)
      & $3.4\times10^{-7}$
      & $3.0\times10^{6}$ \\
    ResNet  (MAC)
      & $3.3\times10^{-7}$
      & $3.0\times10^{6}$ \\
    \midrule
    Static, dense (ours)
      & $4.6\times10^{-8}$ & $2.2\times10^{7}$ \\
    Static, sparse (ours)
      & $4.6\times10^{-8}$ & $2.2\times10^{7}$ \\
    \bottomrule
  \end{tabular}
\end{table}

\subsection{RC circuits for image denoising}
\label{sec:experiment-denoising}

\begin{figure}[t]
  \centering
  \resizebox{0.86\linewidth}{!}{%
  \begin{tikzpicture}[
      font=\small,
      line width=0.55pt,
      state/.style={circle, draw=mediumdraw, fill=white, minimum size=5.5pt,
                    inner sep=0pt},
      gridbox/.style={draw=mediumdraw, fill=mediumfill, rounded corners=4pt,
                      minimum width=1.45cm, minimum height=1.45cm},
      interface/.style={draw, rounded corners=3pt, align=center,
                        minimum height=0.9cm, minimum width=1.8cm}]

    \node[interface, draw=inputdraw, fill=inputfill] (x) at (-3.4,0)
      {noisy\\image $x$};
    \foreach \g/\yc/\lab in {1/2.6/1, 2/0.6/2, 3/-2.4/K} {
      \node[gridbox] (box\g) at (0.45,\yc) {};
      \foreach \i in {0,1,2} {
        \draw[mediumdraw!70] (0,\yc+0.45-0.45*\i) -- (0.9,\yc+0.45-0.45*\i);
        \draw[mediumdraw!70] (0.45*\i,\yc+0.45) -- (0.45*\i,\yc-0.45);
      }
      \foreach \i in {0,1,2} {
        \foreach \k in {0,1,2} {
          \node[state] at (0.45*\k,\yc+0.45-0.45*\i) {};
        }
      }
      \node[above] at (0.45,\yc+0.72) {grid $\lab$};
    }
    \node at (0.45,-0.8) {$\vdots$};

    \draw[->,thick,inputdraw] (x.east) --
      node[above, sloped, pos=0.72, text=black] {$B_1(t)x$} (box1.west);
    \draw[->,thick,inputdraw] (x.east) --
      node[above, sloped, pos=0.72, text=black] {$B_2(t)x$} (box2.west);
    \draw[->,thick,inputdraw] (x.east) --
      node[below, sloped, pos=0.72, text=black] {$B_K(t)x$} (box3.west);

    \node[interface, draw=readdraw, fill=readfill, minimum width=2.9cm]
      (read) at (5.2,0) {pointwise readout};
    \draw[->,thick,readdraw] (box1.east) --
      node[above, sloped, pos=0.5, text=black] {$u_1(T)$} (read.north west);
    \draw[->,thick,readdraw] (box2.east) --
      node[above, sloped, pos=0.5, text=black] {$u_2(T)$} (read.west);
    \draw[->,thick,readdraw] (box3.east) --
      node[below, sloped, pos=0.5, text=black] {$u_K(T)$} (read.south west);
    \node[interface, draw=outdraw, fill=outfill] (out) at (8.6,0)
      {denoised\\image $\hat y$};
    \draw[->,thick,outdraw] (read) -- (out);

  \end{tikzpicture}%
  }
  \caption{RC image-denoising circuit. The noisy image \textcolor{inputdraw}{(blue)} drives
  $K$ parallel nonlinear RC grids \textcolor{mediumdraw}{(green)}.
  The grid states are combined by a pointwise readout \textcolor{readdraw}{(orange)} into the
  denoised image \textcolor{outdraw}{(purple)}.}
  \label{fig:denoising-circuit}
\end{figure}
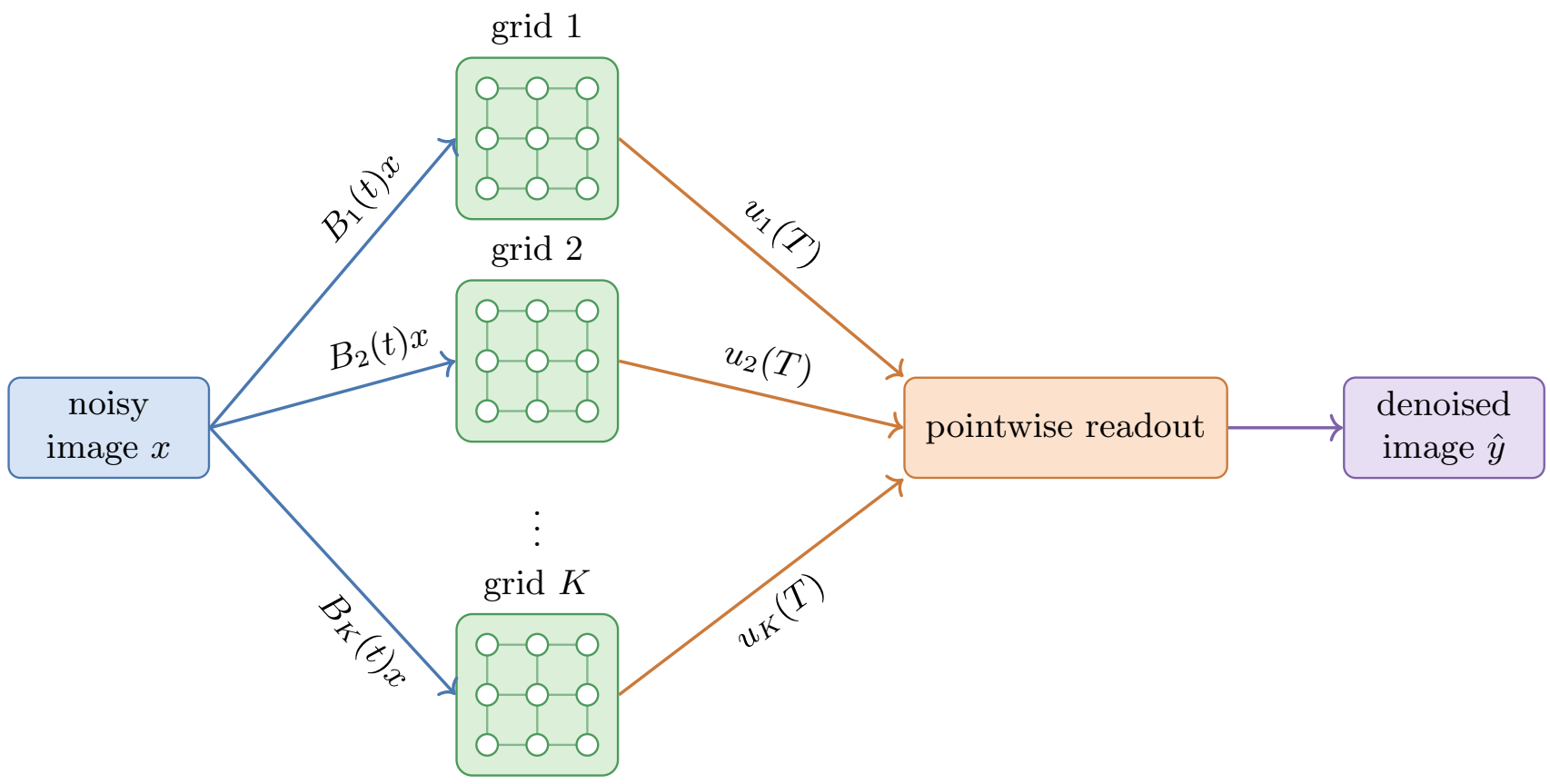

We evaluate grayscale image denoising using the Berkeley
Segmentation Dataset (BSD) \cite{martin2001database}. We use BSD400 for training, and
BSD68 benchmark for testing. Additive Gaussian noise with
standard deviation $\sigma=25/255$ is applied to the clean
images. Training uses randomly sampled $64\times64$ patches.
The circuit consists of $K$ parallel nonlinear RC grids,
each having one node per pixel and resistive connections between
horizontal and vertical neighbors
(\cref{fig:denoising-circuit}). Each node has unit capacitance
and a positive conductance $G_j$ to ground. The dynamics of
grid $j$ are
\begin{equation}
  \frac{du_j}{dt}
  =-D^T\Phi_j(Du_j)+B_j(t)x-G_ju_j,
  \qquad u_j(0)=x,
  \label{eq:denoising-channels}
\end{equation}
where $D$ is the incidence matrix of the pixel lattice and
$B_j(t)x$ represents the injected currents. 
The nonlinear resistive edges follow a smooth, monotone,
saturating current--voltage law, with separate learned
parameters for horizontal and vertical connections.
This produces edge-preserving diffusion, closely related
to total-variation denoising~\cite{rudin1992nonlinear}.

\begin{figure}[t]
  \centering
  \includegraphics[width=\linewidth]{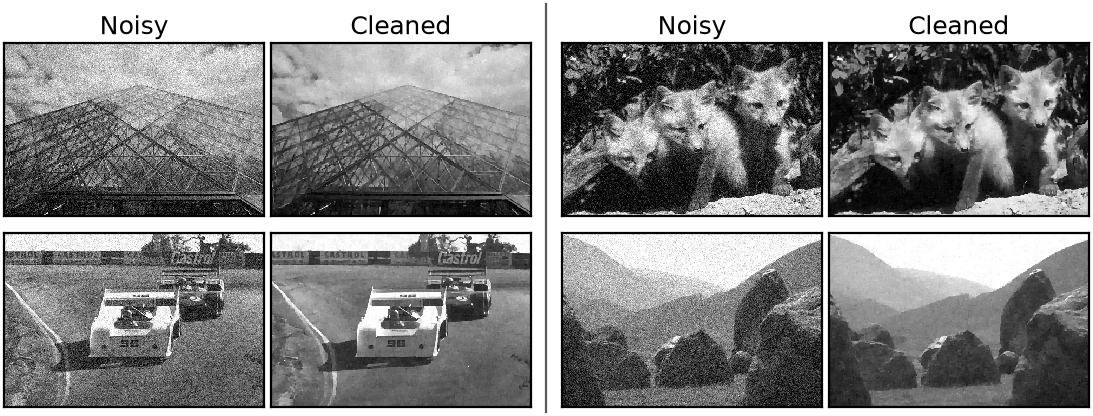}
  \caption{Denoising examples from BSD68 at $\sigma=25\, /255$.
  Each pair shows the noisy input and the output of the
  RC circuit. The learned nonlinear diffusion smooths the noise while
  preserving edges.}
  \label{fig:denoising-examples}
\end{figure}

The input gains $B_j(t)$ are learned over eight time intervals,
and combined through a learned
pointwise affine readout to obtain the denoised image.
We use $K=48$ grids. We report PSNR and SSIM, averaged over BSD68, in
\cref{tab:denoising-results}. Our
circuit attains a PSNR of 27.73\,dB and an SSIM of 0.757, compared with
27.77\,dB and 0.756 for the residual CNN. Hence, on this benchmark, a passive
diffusion device with a pointwise readout reaches a denoising quality
comparable to a trained convolutional network. The examples in
\cref{fig:denoising-examples} show the mechanism behind this result. The
circuit removes the noise in nearly uniform regions, such as the sky and the
hillsides, while it retains edges, such as the frame of the glass pyramid and
the outlines of the rocks. This behavior corresponds to the edge-preserving
diffusion of the learned branch laws.

\begin{table}[t]
  \caption{Image denoising on BSD68 at $\sigma=25/255$. Our nonlinear RC
  circuit with 48 grids attains a PSNR within 0.04\,dB of the residual CNN and
  a higher SSIM, using passive diffusion and a pointwise readout.}
  \label{tab:denoising-results}
  \centering
  \small
  \renewcommand{\arraystretch}{1.2}
  \begin{tabular*}{\linewidth}{@{\extracolsep{\fill}}lcc@{}}
    \toprule
    Model & PSNR (dB)\,\rev{$\uparrow$} & SSIM\,\rev{$\uparrow$}\\
    \midrule
    Residual CNN & 27.77 & 0.756\\
    \midrule
    Nonlinear RC, 48 grids (ours) & 27.73 & 0.757\\
    \bottomrule
  \end{tabular*}
\end{table}

We estimate energy and latency using the assumptions of
\cref{sec:experiment-setup}. For digital interfaces, source
generation and readout require approximately 430 MACs and
96 conversions per pixel, giving an estimated interface energy
of $6.5\times10^{-9}\,\mathrm J$ per pixel. With analog source
gains and current summation, only input and output conversions
are required, reducing this estimate to
$4.0\times10^{-11}\,\mathrm J$ per pixel. For comparison,
the CNN requires 630 MACs per pixel, corresponding to
$6.7\times10^{-9}\,\mathrm J$ under the same digital energy
budget.

Assuming $10^{-15}\,\mathrm F$ capacitance per node and
scaling the largest conductance to $100\,\mu\mathrm S$,
the circuit observation time corresponds to
$2.9\times10^{-10}\,\mathrm s$ per image, independent
of image size due to parallel operation.
\Cref{tab:denoising-energy} summarizes these estimates
alongside the GPU measurements. The analog interface
budget is approximately 170 times smaller than the CNN
MAC reference.
As for the classification circuits, the circuit values are model-based
budgets computed term by term. 
The CNN MAC reference applies the same model to the convolutional network.
Hence, it provides a comparison with the circuit on the same basis.

\begin{table}[t]
  \caption{Energy per denoised pixel and energy efficiency for image
  denoising. The GPU measurement uses an NVIDIA RTX 4090 in float32
  with batch size one. The MAC-model value is an energy estimate.
  The RC values are model-based budgets for 48 grids with digital or
  analog interfaces; they exclude the uncalibrated medium energy
  $\mathcal E_{\mathrm{medium}}$.}
  \label{tab:denoising-energy}
  \centering
  \begin{tabular}{lrr}
    \toprule
    Model & Energy (J/pixel) & Efficiency (pixels/J) \\
    \midrule
    Residual CNN 
      & $1.7\times10^{-7}$ & $5.9\times10^{6}$ \\
    Residual CNN (MAC)
      & $6.7\times10^{-9}$ & $1.5\times10^{8}$ \\
    \midrule
    RC, digital (ours)
      & $6.5\times10^{-9}$ & $1.5\times10^{8}$ \\
    RC, analog (ours)
      & $4.0\times10^{-11}$ & $2.5\times10^{10}$ \\
    \bottomrule
  \end{tabular}
\end{table}

\subsection{Inductive circuits for language modeling}
\label{sec:experiment-language}

The third experiment is a teacher--student study of character-level language
modeling on Tiny Shakespeare. The vocabulary contains 65 characters, and all
models use a context of 64 characters. We compare two passive students, having
three layers of either 512 or 1024 nodes, with two transformer references that
we denote by T4 and T6. T4 has four transformer layers of width 128 and
804{,}096 parameters. T6 has six transformer layers of width 384 and
10{,}671{,}360 parameters. Both transformers are evaluated at context 64.

The teacher is T6. Rather than exposing the student only to the characters in
the original corpus, we first use the frozen T6 model to generate training
text. T6 samples each new character at temperature one from its distribution
conditioned on the preceding 64 characters. For every generated character, we
store both the sampled character and the complete 65-dimensional log
probability vector from which it was drawn. The student therefore receives not
only a single target character but also the teacher's relative probabilities
for all possible next characters. The real Tiny Shakespeare validation split
is kept separate and is used to compare the students with T4 and T6.

\begin{figure}[t]
  \centering
  \resizebox{0.96\linewidth}{!}{%
  \begin{tikzpicture}[
      font=\small,
      line width=0.55pt,
      state/.style={circle, draw=mediumdraw, fill=white,
                    minimum size=6pt, inner sep=0pt},
      layer/.style={draw=mediumdraw, fill=mediumfill, rounded corners=4pt,
                    minimum width=0.8cm, minimum height=3.5cm},
      interface/.style={draw, rounded corners=3pt, align=center,
                        minimum height=0.9cm, minimum width=1.7cm},
      branchlabel/.style={below, align=center, font=\scriptsize,
                          text=black!70}]

    \node[layer] (lay0) at (0,0) {};
    \node[layer, draw=centerdraw, fill=centerfill] (lay1) at (3.8,0) {};
    \node[layer] (lay2) at (7.6,0) {};

    \foreach \j/\yy in {1/1.35,2/0.45,3/-0.45,4/-1.35} {
      \node[state] (left\j)  at (0,\yy) {};
      \node[state, draw=centerdraw] (mid\j) at (3.8,\yy) {};
      \node[state] (right\j) at (7.6,\yy) {};
    }

    \foreach \a in {1,2,3,4} {
      \foreach \b in {1,2,3,4} {
        \draw[mediumdraw!45] (left\a) -- (mid\b);
        \draw[mediumdraw!45] (mid\a) -- (right\b);
      }
    }

    \foreach \j in {1,2,3,4} {
      \node[state] at (left\j) {};
      \node[state, draw=centerdraw] at (mid\j) {};
      \node[state] at (right\j) {};
    }

    \node[above, align=center] at (0,1.85) {outer layer\\$N$ nodes};
    \node[above, align=center] at (3.8,1.85) {middle layer\\$N$ nodes};
    \node[above, align=center] at (7.6,1.85) {outer layer\\$N$ nodes};

    \node[interface, draw=inputdraw, fill=inputfill] (eone) at (-3.1,0)
      {embedding\\$E_1(x_k)$};
    \node[interface, draw=inputdraw, fill=inputfill] (etwo) at (10.7,0)
      {embedding\\$E_2(x_k)$};

    \draw[->, thick, inputdraw] (eone.east) --
      node[above, font=\scriptsize, text=black] {excitation} (lay0.west);
    \draw[->, thick, inputdraw] (etwo.west) --
      node[above, font=\scriptsize, text=black] {excitation} (lay2.east);

    \node[branchlabel] at (1.9,-1.85)
      {$N^2$ nonlinear\\inductive branches};
    \node[branchlabel] at (5.7,-1.85)
      {$N^2$ nonlinear\\inductive branches};

    \node[interface, draw=readdraw, fill=readfill, minimum width=2.4cm]
      (head) at (3.8,-4.5) {normalization and\\linear readout};
    \node[interface, draw=outdraw, fill=outfill, minimum width=2.0cm]
      (logits) at (3.8,-6.25) {logits for $x_{k+1}$};

    \draw[->, thick, readdraw] (lay1.south) --
      node[right, pos=0.65, inner sep=3pt, text=black] {$u_{\mathrm{mid}}$}
      (head.north);
    \draw[->, thick, outdraw] (head.south) -- (logits.north);

  \end{tikzpicture}%
  }
  \caption{Two-sided three-layer inductive language model. Independent
  embeddings of character $x_k$ \textcolor{inputdraw}{(blue)} excite the two outer
  layers \textcolor{mediumdraw}{(green)}. Complete bipartite inductive interfaces transmit both
  responses toward the middle layer \textcolor{centerdraw}{(yellow)}, whose state carries the
  history of earlier characters and is decoded by the readout \textcolor{readdraw}{(orange)}
  into next-character logits \textcolor{outdraw}{(purple)}. Only four of the $N$ nodes in
  each layer are drawn.}
  \label{fig:two-sided-language-medium}
\end{figure}
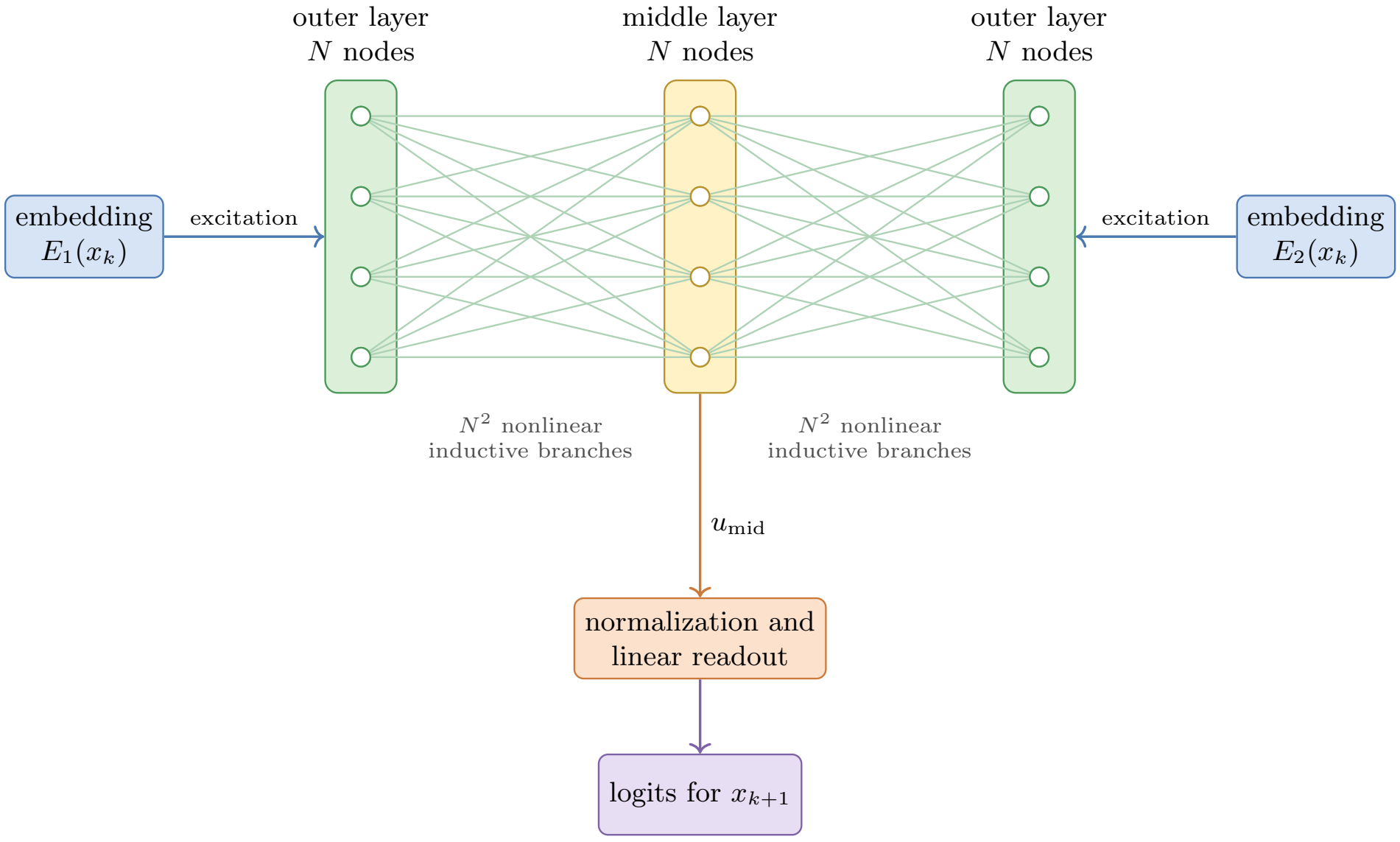

Each student is a three-layer nonlinear inductive medium. All three layers
have width $N$, with $N=512$ or $N=1024$. Consecutive layers are connected by
complete bipartite graphs, giving $N^2$ nonlinear inductive branches between
the first and middle layers and another $N^2$ branches between the middle and
last layers. A character is encoded independently at the two outer layers.
The two excitations propagate inward from opposite sides and interact through
the state of the middle layer, which is used for the readout. Thus, the
architecture is both two-sided and causal: the outer layers receive the
current character, while the middle-layer state used for prediction contains
only information that has already propagated from earlier characters,
as we illustrate in \cref{fig:two-sided-language-medium}.
Together, the two input banks and shared middle bank form the complete
bipartite graph $K_{2N,N}$. The shared bank aggregates both inputs with
retained history, loosely resembling attention's aggregation of
representations~\cite{vaswani2017attention}; here the interactions use
learned physical couplings rather than query--key attention weights.

In normalized variables, the dynamics are a three-layer instance of
\eqref{eq:nonlinear-wave-restoring}. The branches use the flux--current law in \cref{eq:tanh-inductor},
with its inverse supplying the restoring current; the node capacitances supply the second-order state,
and learned positive damping controls the memory time. A small grounding term
removes the unconstrained common mode. We use a damped leapfrog update with
three time steps per character. Time-step refinement gives a relative error
below $3\%$. The state is
retained across the 65-character training window, so language history is
stored in the displacement and velocity of the physical medium rather than in
an explicit digital recurrent state.

Let $p_{T6,k}$ be the distribution stored from T6 and let $p_{S,k}$ be the
student distribution at position $k$. The prediction loss in \eqref{eq:physical-design-problem} is
knowledge distillation,
\begin{equation}
  \mathcal L_{\mathrm{distill}}
  =\frac{1}{K}\sum_{k=1}^{K}
    \operatorname{KL}\!\left(p_{T6,k}\,\|\,p_{S,k}\right),
  \label{eq:language-distillation}
\end{equation}
with the blind initial prediction omitted. This teacher--student formulation
is useful for the physical model because the full teacher distribution gives
a training signal for every output character, including alternatives that
were not sampled. Cross-entropy is evaluated in nats per character on real
validation text. 

For a width $N$, the student contains $2N^2$ trained nonlinear inductive
couplings, $3N$ damping parameters, two $65\times N$ character embeddings,
and a layer-normalized linear readout from the middle layer to 65 logits. Its
total parameter count is therefore
\begin{equation}
  2N^2+200N+65.
  \label{eq:language-parameter-count}
\end{equation}
This gives 626{,}753 parameters for the $3\times512$ student and
2{,}302{,}017 parameters for the $3\times1024$ student. The smaller physical
student has fewer parameters than T4, whereas the larger student has about
$2.9$ times as many parameters as T4 but only about $22\%$ as many as T6.
Unlike the transformer parameters, most student parameters are physical branch
values: the two bipartite interfaces alone account for $2N^2$ parameters.

T4 is not used to train the passive models; it provides a compact transformer
baseline. T6 serves both as the larger comparison model and as the teacher.
On the fixed Tiny Shakespeare validation probe, their reference
cross-entropies are 1.855 nats per character for T4 and 1.345 for T6. The same
validation windows and character positions are used for both passive students.

\begin{table}[t]
  \caption{Model sizes and transformer validation results on Tiny Shakespeare.
  T4 and T6 have context 64; T6 is the teacher. Passive students have
  three physical layers and two-sided input. Lower cross-entropy and BPC
  are better.}
  \label{tab:language-results}
  \centering
  \small
  \setlength{\tabcolsep}{3pt}
  \renewcommand{\arraystretch}{1.2}
  \begin{tabularx}{\linewidth}{@{}Xrccc@{}}
    \toprule
    Model & Parameters & Context &
      \shortstack{Validation CE\\(nats/character)} & BPC\\
    \midrule
    T4: 4-layer, width 128 transformer
      & 804{,}096 & 64 & 1.855 & 2.677\\
    T6: 6-layer, width 384 transformer
      & 10{,}671{,}360 & 64 & 1.345 & 1.940\\
    \midrule
    Passive student, $3\times512$ (ours)
      & 626{,}753 & 64 & 1.471 & 2.1312\\
    Passive student, $3\times1024$ (ours)
      & 2{,}302{,}017 & 64 & 1.397 & 2.0152 \\
    \bottomrule
  \end{tabularx}
\end{table}

The energy accounting separates the medium, input interface, sensing, and
readout:
\begin{equation}
  \mathcal E_{\mathrm{character}}
  =\mathcal E_{\mathrm{medium}}+
    \mathcal E_{\mathrm{input}}+
    \mathcal E_{\mathrm{sense}}+
    \mathcal E_{\mathrm{readout}}.
  \label{eq:language-energy-components}
\end{equation}
The readout is an implementation choice, not a fixed energy cost of the
medium. \Cref{tab:language-energy} gives two reference electronics budgets
for the stated normalization and linear projection. The $2N$ character
embedding values can be precomputed and read from a lookup table, with no
encoding MACs. For eight-bit stored values, their packed lookup cost is
$(N/2)e_{\mathrm r}$. Input conversion and buffering cost
$2N(e_{\mathrm c}+e_{\mathrm b})$, and sensing the $N$ middle-layer values
costs $N(e_{\mathrm c}+e_{\mathrm b})$.

A digital readout requires $65N$ MACs and weight reads; we apply
\eqref{eq:digital-energy-budget}, including its allowance for normalization,
probability processing, and control. For an alternative analog matrix
projection, the weights remain in local analog memory. We use the measured
eight-bit matrix--vector energy of Le Gallo et al.~\cite{legallo2023mixedsignal}
as a reference, accounting for padding the $65\times N$ projection to
$\lceil N/256\rceil$ active $256\times256$ tiles. The projection budget is
\begin{equation}
  \widehat{\mathcal E}_{\mathrm{proj}}
  =2(256)^2\lceil N/256\rceil e_{\mathrm{op}},\qquad
  e_{\mathrm{op}}\in[1.02,4.03]\times10^{-13}\,\mathrm J.
  \label{eq:analog-readout-budget}
\end{equation}
The factor two counts a MAC as two operations. This reference includes the
matrix engine's conversion and peripheral costs; final output conversion
is therefore not added a second time. We retain the separate middle-layer
sensing cost for normalization and allocate
$N(5\times10^{-11}\,\mathrm J)$ to normalization, probability processing, and
readout control. The range spans the reference chip's one- and four-phase
read modes; it is not an uncertainty interval for a built student device.
Other analog, digital, or mixed-signal readouts can be substituted in
\eqref{eq:language-energy-components}.

The student entries in \cref{tab:language-energy} are electronics subtotals:
the total additionally includes $\mathcal E_{\mathrm{medium}}$. We do not
assign that term a value in joules until the normalized dynamics, signal
amplitudes, drive waveform, and reset are physically calibrated. In
particular, assigning $10^{-15}\,\mathrm F$ per node alone does not determine
it.

For the transformer reference, incremental inference with a reusable cached
context of length $H=64$ requires
$n_{\mathrm{MAC}}=L(12d^2+2Hd)+65d$, where $L$ is the number of layers and
$d$ their width. Counting one main local weight or key/value read per MAC
gives 860{,}288 MACs for T4 and 10{,}936{,}704 for T6. With
\eqref{eq:digital-energy-budget}, their active-electronics budgets are
$9.1\times10^{-6}$ and $1.2\times10^{-4}\,\mathrm J$ per character.
These are optimized cached-inference scenarios with on-chip storage,
not measurements of the current generation code or of a GPU.

\begin{table}[t]
  \caption{Language-model energy components, in J/character. Student
  electronics budgets depend on the input, sensing, and readout
  implementations. Digital and analog-matrix readouts are alternatives;
  their costs are not added together. Each student subtotal must be added
  to its uncalibrated medium energy to obtain a total. Analog ranges span
  the two reference read modes. Transformer rows give the separate
  model-based digital reference.}
  \label{tab:language-energy}
  \centering
  \small
  \setlength{\tabcolsep}{3pt}
  \renewcommand{\arraystretch}{1.2}
  \begin{tabularx}{\linewidth}{@{}>{\raggedright\arraybackslash}Xcc@{}}
    \toprule
    Component or implementation & $3\times512$ & $3\times1024$\\
    \midrule
    Input lookup, conversion, and buffering
      & $2.2\times10^{-8}$ & $4.4\times10^{-8}$\\
    Middle-layer sensing
      & $1.0\times10^{-8}$ & $2.0\times10^{-8}$\\
    Physical medium
      & $\mathcal E_{\mathrm{medium},512}$ & $\mathcal E_{\mathrm{medium},1024}$\\
    Digital readout, including normalization and control
      & $3.5\times10^{-7}$ & $7.0\times10^{-7}$\\
    Analog matrix readout, including normalization and control
      & $[5.2\times10^{-8},\,1.3\times10^{-7}]$
      & $[1.0\times10^{-7},\,2.6\times10^{-7}]$\\
    \midrule
    Electronics subtotal: digital readout
      & $3.8\times10^{-7}$ & $7.7\times10^{-7}$\\
    Electronics subtotal: analog matrix readout
      & $[8.4\times10^{-8},\,1.6\times10^{-7}]$
      & $[1.7\times10^{-7},\,3.3\times10^{-7}]$\\
    \midrule
    T4, MAC reference & \multicolumn{2}{c}{$9.1\times10^{-6}$}\\
    T6, MAC reference & \multicolumn{2}{c}{$1.2\times10^{-4}$}\\
    \bottomrule
  \end{tabularx}
\end{table}

\section{Conclusions and outlook}
\label{sec:conclusions}

We have presented a framework for designing passive physical systems as
computational devices. The central premise is that a computation need not be
implemented as a prescribed sequence of digital arithmetic operations. It can
instead be represented by the equilibrium or transient response of a physical
medium. In this view, the graph, component laws, boundary excitations, and
measurement procedure jointly define the computational architecture, while
training selects physically admissible component parameters. Passivity is not
added as a penalty after training: it is enforced by the circuit construction
and retained for every admissible parameter value.

The three experiments illustrate why no single physical architecture should be
expected to work best for every task. Static classification uses a layered
bipartite resistive network, with data entering at one boundary and class
logits read at the other. Image denoising uses several nonlinear RC diffusion
channels on the image grid. Their untrained dynamics already provide a useful
smoothing prior, while learning adapts the anisotropy, time dependence, and
combination of the diffusion responses. Character-level language modeling uses
a three-layer inductive medium excited from both sides. Its evolving physical
state supplies memory, and the middle layer combines the two transmitted
responses. The teacher--student construction further shows how a large digital
model can transfer its output distribution to a substantially smaller passive
student without prescribing how the student must perform the computation.

These examples also clarify the source of the potential energy advantage.
Dense coupling is expensive to simulate because it requires many arithmetic
operations and repeated movement of parameters and states. In a physical
device, the same interactions occur concurrently through the connected
components. The potential saving must nevertheless be assessed for the complete
implementation. The corrected budgets account separately for the input
interface, sensing, and readout, and show how alternative readout hardware
changes these contributions. The normalized wave models still require physical
calibration before their total energy can be quantified. Architecture selection
must therefore consider accuracy, component count, wiring, response time, and
interface energy together.

Several limitations remain. The results in this work are obtained from
numerical models, and the component values and normalized time scales have not
yet been calibrated to a single fabrication process. The model-based energy entries are
engineering budgets rather than measurements; the student electronics
subtotals exclude the uncalibrated medium contribution.
Hardware validation would require selecting and characterizing components and
testing how component tolerances, thermal noise, and sensing precision affect
performance. We leave that validation to future work.

Training the simulated device remains a
digital optimization problem whose cost can be high, although that cost is
amortized when the trained device is evaluated repeatedly.

A broader challenge is the systematic selection of physical architectures.
The examples here were chosen using insight about the task: equilibrium for
static classification, diffusion for spatial denoising, and waves for
sequential memory. Future work should turn this choice into a design problem
over graph topology, physical laws, placement of sources and sensors, and
operating time. Sparse, local, multiscale, and modular networks may provide
better compromises between expressiveness and manufacturability than dense
graphs. Hybrid systems are another promising direction: a passive medium can
perform the expensive spatial or temporal transformation, while a small
digital circuit handles encoding, calibration, and task-specific readout.

Ultimately, nature processing units invite a form of hardware-algorithm
co-design in which the governing equations are part of the model architecture.
The aim is not to reproduce every operation of a conventional neural network
in analog hardware. It is to identify physical dynamics whose native behavior
already resembles the desired computation, and then learn the device that
turns this behavior into a useful input-output map. Establishing when this
approach offers a reliable advantage will require calibrated hardware,
reproducible energy measurements, and comparisons against efficient digital
implementations. The results developed here provide a mathematical and
experimental starting point for that program.

\bibliographystyle{siamplain}
\bibliography{references}

\end{document}